\documentclass[11pt,a4paper]{article}
\usepackage[hyperref]{emnlp2020}
\usepackage{times}
\usepackage{latexsym}
\usepackage{subfigure}
\usepackage{makecell}
\usepackage{arydshln}
\usepackage{amsfonts}
\usepackage{enumitem}

\usepackage{graphicx}

\usepackage{multirow}
\usepackage{microtype}

\aclfinalcopy 

\title{Meta-learning for Few-shot Natural Language Processing: A Survey}

\author{Wenpeng Yin \\
  Salesforce Research \\
  \texttt{wyin@salesforce.com}}

\date{}

\begin{document}
\maketitle
\begin{abstract}

Few-shot natural language processing (NLP) refers to  NLP tasks that are accompanied with merely a handful of labeled examples.   This is a real-world challenge that an AI system must learn to handle.  Usually we rely on collecting more auxiliary information or developing a more efficient learning algorithm. However, the general gradient-based optimization in high capacity models, if training from scratch,  requires many parameter-updating  steps over a large number of labeled examples to perform well \cite{DBLPSnellSZ17}.


If the target task itself cannot provide more information, how about collecting more tasks equipped with rich annotations to help the model learning?  The goal of meta-learning is to train a model on a variety of tasks with rich annotations, such that it can solve a new  task using only a few  labeled samples. The key idea  is to train the model's initial parameters such that the model has maximal performance on a new task after the parameters have been updated through zero or a couple of gradient steps.

There are already some surveys for meta-learning, such as \cite{DBLPVilaltaD02,DBLPabs03548,DBLP05439}. Nevertheless,  this paper focuses on NLP domain, especially few-shot applications. We try to provide  clearer definitions, progress summary and some common datasets  of applying meta-learning to few-shot NLP.
\end{abstract}

 


\section{What is meta-learning?}
To solve a new task which has only a few examples,  meta-learning aims to build \textbf{efficient}  algorithms (e.g., need a few or even no task-specific fine-tuning) that can learn the new task quickly.

Conventionally, we train a task-specific model by iterating on the task-specific labeled examples. For example, we treat an input sentence as a training example in text classification problems. In contrast, the meta-learning framework \textbf{treats  tasks as training examples}---to solve a new task, we first collect lots of tasks, treating each as a training example and train a model to adapt to all those training tasks, finally this model is expected to work well for the new task. 

In the regular text classification tasks, we usually assume that the training sentences and test sentences are in the same distribution. Similarly,  meta-learning also assumes that the training tasks and the new task are from the same distribution of tasks $p(\mathcal{T})$. During
meta-training, a task $T_i$ is sampled from $p(\mathcal{T})$, the model is trained with $K$ samples, and then tested on test set 
from $T_i$. \textit{The test error on the sampled task  $T_i$ serves as the training error of the meta-learning process at the current $i$-th iteration}\footnote{Here the ``test error'' is the training loss, because what we really care is the test performance on the target task.}. After the meta-training, the new task, sampled from $p(\mathcal{T})$ as well, measures  the model’s performance after learning from $K$ samples.

Since the new task only has $K$ labeled examples and a large set of unlabeled test instances, each training task also keeps merely $K$ labeled examples during the training. This is to make sure that the training examples (means those training tasks here) have the same distribution as the test example (means the new task here). Usually, the $K$ labeled examples are called ``support set''.

To describe meta-learning at a higher level: meta-learning doesn't learn how to solve a specific task. It successively learns to solve many tasks. Each time it learns a new task, it becomes better at learning new tasks: it learns to learn if ``its performance at each task improves with experience and with the number of tasks'' \cite{DBLPThrunP98}.

\paragraph{Meta-learning vs. Transfer learning.} Conventionally, transfer learning uses past experience of a source task to improve learning on a target task --- by pre-training a parameter prior plus optional fine-tuning. Transfer learning refers to a problem area (task A helps task B) while meta-learning refers to methodology which can be used to improve transfer learning as well as other problems \cite{DBLP05439}. 

Technically, the pretraining in transfer learning often does not take its ultimate application scenario (e.g., a few-shot task) into consideration; meta-learning, instead, is optimized particularly towards benefiting the target task (for example, the system configuration is optimized so that it only needs a few gradient updates in a target few-shot problem). 

In effect, meta-learning assumes that the training tasks are in the same distribution with the target task; this often means that all the so called tasks (including training tasks and the target task) are essentially the same problem in different domains, such as from reviews of other product domains to the target cellphone's review domain. Transfer learning, instead, does not have such a strict assumption; in theory, transfer learning can pretrain on any source tasks that can be potentially helpful to the target task (such as from a question answering task to a coreference resolution task).

\begin{figure}[t]
\centering
\includegraphics[width=7cm]{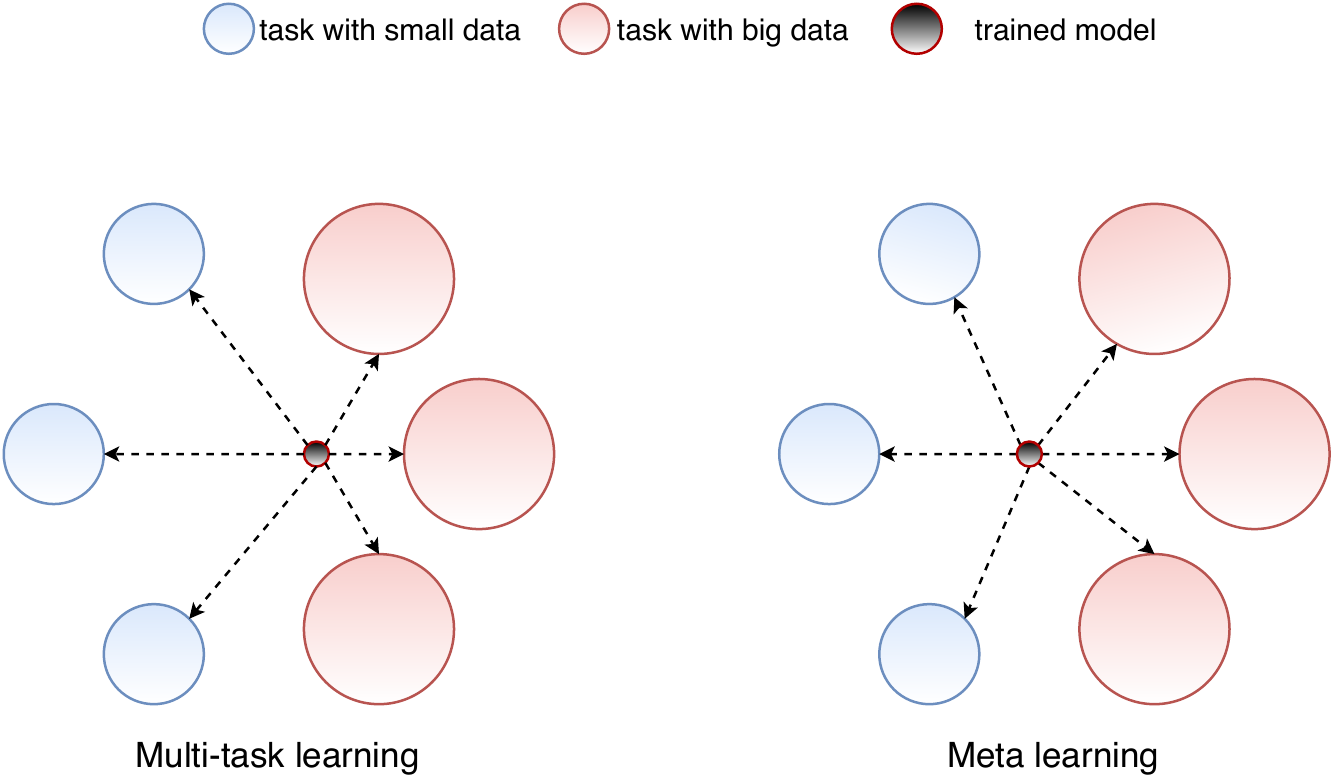}
\caption{Multitask learning vs. meta learning. This figure is adapted from \newcite{DBLPDouYA19}.} \label{fig:metamulti}
\end{figure}

\paragraph{Meta-learning vs. Multi-task learning.} If we think the aforementioned transfer learning is often implemented as a sequential training flow from source tasks to the target task, multi-task learning is to train all the tasks together simultaneously. 

Since meta-learning also relies on a set of training tasks, meta-learning is also a kind of multi-task learning. We summarize three differences here:
\begin{itemize}
    \item The conventional goal of multi-task learning  is to learn a well pretrained model that can generalize to the target task; meta-learning tries to learn an efficient learning algorithm that learns the target task quickly. 
    \item In addition, multi-task learning may favor tasks with significantly larger amounts of data than others, as shown in Figure \ref{fig:metamulti}.
    \item Since meta-learning treats tasks as training examples, ideally, the more training tasks the better. However, multi-task learning may meet increasing challenges in training simultaneously over too many tasks. 
\end{itemize}

\section{Meta-learning milestones}\label{sec:intro}
Transferable knowledge in meta-learning is derived in the form of generalizable representation space  or optimization strategies. The target few-shot task is then  handled in a feed-forward distance function without updating network weights or learned by fine-tuning with the efficient optimization strategy.

\begin{table*}[t]
  \centering
  \begin{tabular}{c|c|c|c}
  name & embedding function & similarity metric  & other notes\\\hline
 
 Siamese Network  & \multirow{2}{*}{deep neural net} & sigmoid over  &  \\
\cite{koch2015siamese} &  &weighted $L_1$ distance & \\\hline

Matching Network  & \multirow{2}{*}{deep neural net} & \multirow{2}{*}{cosine} & the embedding function \\
\cite{DBLPVinyalsBLKW16} &  & & depends on the support set\\\hline

Prototypical Network  & \multirow{2}{*}{deep neural net} & \multirow{2}{*}{Euclidean distance} & compare the test example with the\\
\cite{DBLPSnellSZ17} &  & &   classes rather than support examples\\\hline

Relation Network  & \multirow{2}{*}{deep neural net} & deep  net  & \\
\cite{DBLPSungYZXTH18} &  &outputs one scalar &   \\\hline


\end{tabular}
\caption{Some representative metric-learning literature.}\label{tab:metriclearning}
\end{table*}

\subsection{Learning to embed: metric-based meta-learning}
Metric-based meta-learning (or called ``metric learning'') learns a distance function between data points so that it classifies  test instances by comparing them  to the $K$ labeled examples. The ``distance function'' often consists of two parts: one is an embedding function which encodes any instances into a representation space, the other is a similarity metric, such as cosine similarity or Euclidean distance,  to calculate how close two instances are in the space. If the distance function is learnt well on the training tasks, it can work well on the target task without fine-tuning.

\paragraph{Siamese Network.} \newcite{koch2015siamese} proposed a Siamese network which takes two instances as input and outputs a scalar indicating they  belong to the same class or not. The Siamese network, trained on training tasks, is essentially a distance function. However, it does not follow the principle of meta-learning: Siamese network was neither trained specifically to minimize the test losses on training tasks nor trained to learn an efficient gradient-based algorithm. 

\paragraph{Matching Network.} \newcite{DBLPVinyalsBLKW16} proposed ``matching network'', the  first metric-based meta-learning algorithm, to solve one-shot problem. Matching network is essentially a parametric nearest neighbors algorithm, defined as follows:
\begin{equation}
    P(\hat{y}|\hat{x}, S) = \sum^k_{i=1} a(\hat{x}, x_i) y_i
\end{equation}
where $S$ is a support set containing $k$ labeled examples \{$x_i$, $y_i$\}, $i=1,\cdots, k$; $\hat{x}$ is a text example with its gold label $\hat{y}$. $a(\cdot)$ is a similarity function given the representations of $\hat{x}$ and $x_i$. One contribution is that each support/testing example learns its representation with the background $S$; it means the whole support set influences the representation learning of each example. In experiment, they tried one-shot problems with and without fine-tuning; but fine-tuning does not show improvements. Matching networks can be interpreted as a weighted nearest-neighbor classifier applied within an embedding space \cite{DBLPSnellSZ17}.

\paragraph{Prototypical Network.} \newcite{DBLPSnellSZ17} proposed ``Prototypical Networks'' with two novelties compared with ``Matching Networks'': (i) Using class representations rather than example representations in the support set; (ii) They found the choice of similarity metric is vital---Euclidean distance outperforms cosine similarity. Prototypical Networks differ from Matching Networks in the few-shot case with equivalence in the one-shot scenario. In addition, This literature  claimed that the support-set-aware representation learning is unnecessary.

\paragraph{Relation Network.} \newcite{DBLPSungYZXTH18} proposed ``Relation Network'' that defines the metric as:
\begin{equation}
    r_{i,j}=s(e(x_i), e(x_j))
\end{equation}
where function $e()$ is an embedding function which generates a representation vector for any input instance; function $s()$ is a scoring function that produces a scalar $r_{i,j}$ between 0 and 1 representing the similarity between the two elements $x_i$ and $x_j$. Different with the ``Prototypical Networks'', the scoring function $s()$ is a deep neural network rather than the Euclidean distance.

Following the routine of metric-based meta learning, \newcite{DBLPuGYCPCTWZ18}  learned multiple metrics for few-shot text classification problems.
Essentially, metric-learning is  a pretrained nearest-neighbor algorithm.

\begin{figure}[t]
\centering
\includegraphics[width=7cm]{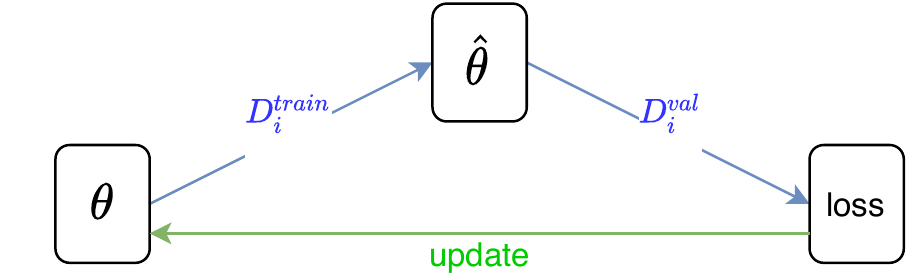}
\caption{MAML meta-learning} \label{fig:maml}
\end{figure}
\begin{figure}[t]
\centering
\includegraphics[width=7cm]{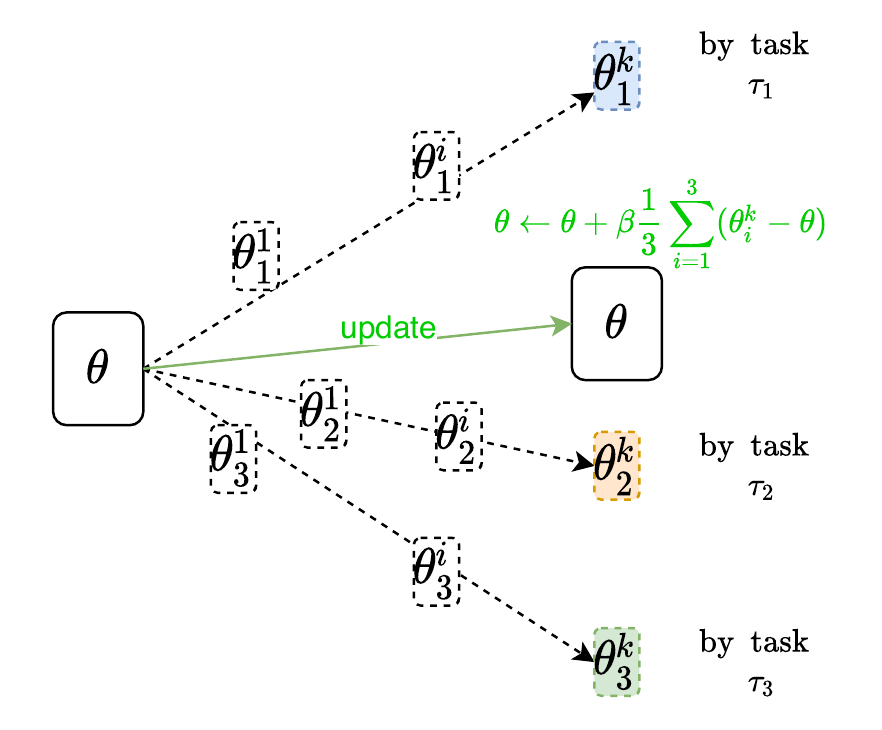}
\caption{Reptile meta-learning (batched version)} \label{fig:reptile}
\end{figure}
\subsection{Learning to fine-tune: optimization-based meta-learning}
Optimization-based methods learn a good point of parameter initialization  for a neural model from which a few steps of gradient descent, given a few examples, can   reach the optimal point for a new task. For each training task (it has $train$ and $validation$\footnote{They correspond to the $support$ and $test$ of a target task.}), the rationale is ``how to fine-tune the model on $train$ so that it can perform well on $validation$''. In order to get good performance on the $validation$ of each training task, the meta-learning uses the validation error on $validation$ as the optimization loss. This loss is implemented through a two-step procedure: first assume the model has fine-tuned on the $train$, obtaining the updated parameters (here are just some ``assumed'' updated parameters, the original model parameters has not been updated in reality), then applying these updated parameters to predict the $validation$, getting the  error which is converted as a loss value; this loss is used to compute gradients, and the original parameters will be updated at this step.

\paragraph{MAML.} \newcite{DBLPFinnAL17} proposed MAML (model-agnostic meta learning) which consists of the following steps in one episode:
\begin{itemize}
\item Create a copy of the model with its initial parameters $\theta$.
\item Train the model on the training set $\mathcal{D}_i^{train}$ (only a few gradient descents):
\begin{equation}
    \hat{\theta} = \theta-\alpha\bigtriangledown_\theta \mathcal{L}_i(\theta, \mathcal{D}_i^{train})
\end{equation}
\item Apply the  model with the updated parameters $\hat{\theta}$ on the validation set $\mathcal{D}_i^{val}$.
\item Use the loss on the validation set to update the initial parameters $\theta$.
\begin{equation}
\label{eq:maml}
    \theta = \theta-\beta\bigtriangledown_\theta\sum_i\mathcal{L}_i(\hat{\theta}, \mathcal{D}_i^{val})
\end{equation}
\end{itemize}
Then, in the next episode,  MAML runs the same process on a newly sampled training task. The process is depicted in Figure \ref{fig:maml}.

During meta-training, the MAML learns initialization parameters that allow the model to adapt quickly and efficiently to a new  task with a few examples.

MAML is model agnostic; this means that it can virtually be applied to any neural networks. However, MAML is quite hard to train because there are two levels of training: the meta-backpropagation implies the computation of gradients of gradients.



\paragraph{FOMAML---First-Order MAML \cite{DBLPFinnAL17}.} The standard MAML has expensive computation. FOMAML is a simplified implementation as follows:
\begin{equation}
\label{eq:fomaml}
    \theta = \theta-\beta\bigtriangledown_{\hat{\theta}}\sum_i\mathcal{L}_i(\hat{\theta}, \mathcal{D}_i^{val})
\end{equation}

Comparing the Equations \ref{eq:maml}-\ref{eq:fomaml}, FOMAML updates the original parameters $\theta$ by considering only the gradients on the last version of ``fake'' parameters $\hat{\theta}$. So, the gradients from $\hat{\theta}$ to the $\theta$ is omitted.

\paragraph{Reptile \cite{DBLP02999}.} Reptile is another first-order optimization-based meta-learning, as shown in Figure \ref{fig:reptile}. It also samples training tasks from $p(\mathcal{T})$: $\tau_1$, $\cdots$, $\tau_i$, $\cdots$, $\tau_n$. Each training task does not have \{$\mathcal{D}_i^{train}$, $\mathcal{D}_i^{val}$\} separations. For  training task $\tau_i$, let's assume the original parameters $\theta$ have  went through $m$ steps of updating and become $\theta_i^m$ (i.e., $\theta_i^m=\mathrm{SGD}(L_{\tau_i}, \theta, m)$), then Reptile updates $\theta$ as follows:
\begin{equation}
\label{eq:reptile}
    \theta = \theta+\beta\frac{1}{n}\sum_{i=1}^n (\theta_i^m-\theta)
\end{equation}

The Reptile algorithm looks like the standard SGD in minibatch; if $m=1$, they are the same; if $m>1$, the expectation $\mathop{\mathbb{E}}_{\tau}[\mathrm{SGD}(L_\tau, \theta, m)]$ differs from $\mathrm{SGD}(\mathop{\mathbb{E}}_{\tau}[L_\tau], \theta, m)$ 

MAML explicitly optimizes the efficiency of the algorithm on the support set, making sure the learnt algorithm can learn fast in the few-shot examples of the target task. In contrast, Reptile tries to optimize the system so that it can work well on all training tasks---it may work well if the target task is very close to the training tasks. 

\begin{table*}[t]
  \centering
  \begin{tabular}{c|l}
  model & \multicolumn{1}{|c}{main notes}\\\hline
  
MAML & \textbullet\enspace Propose the basic framework for optimization-based meta-learning  \\
\cite{DBLPFinnAL17} & \textbullet\enspace Agnostic to model architecture and task specification\\\hline

FOMAML & \multirow{2}{*}{\textbullet\enspace First-order MAML algorithm}  \\
\cite{DBLPFinnAL17} & \\\hline

Reptile & \textbullet\enspace First-order optimization algorithm  \\
\cite{DBLP02999} & \textbullet\enspace Agnostic to model architecture and task specification\\\hline

ATAML & \multirow{2}{*}{\textbullet\enspace Task-agnostic parameters \&  task-specific parameters}  \\
\cite{jiang2018attentive} & \\\hline

 & \textbullet\enspace Meta-train on different tasks  \\
LEOPARD & \textbullet\enspace Similar with \cite{jiang2018attentive}, set task-agnostic parameters  \\
\cite{DBLP03863} & and  task-specific parameters\\
& \textbullet\enspace Use BERT as text encoder\\\hline




\end{tabular}
\caption{Some representative optimization-based meta-learning (some are from  NLP)}\label{tab:testondownstreamNLI}
\end{table*}
\section{Progress specific to few-shot NLP}
Usually, the progress of meta-learning is split by its techniques, such as metric-based or optimization-based. Whichever technique applies, the applications are often limited to simulated datasets where each classification label is treated as a task. To be specific to NLP problems, we separate the progress in the following two categories: (i) Meta-learning  on different domains of the same problem. This categories usually have access of different domains of datasets which essentially belong to the same problem, such as different domains of datasets for sentiment classification, different domains of datasets for intent classification; (ii) Meta-learning on  diverse problems and then it is applied to solve a new problem. 

\subsection{Meta-learning within  a problem}
Here, ``the same problem'' is often studied in two types: one is one dataset of multiple classes where each class is treated as a task; the other is related to exploring the same task, such as sentiment classification, on different domains. 

\paragraph{A class is a task.} 
\newcite{jiang2018attentive}  studied topic classification problem in which each topic is a task. Some topics are training tasks, some are testing tasks. ATAML, an extension of the MAML \cite{DBLPFinnAL17} algorithm was proposed by grouping the parameters into task-agnostic part and task-specific part. It also has two levels of optimizations: at the task-level, only the task-specific parameters will be calculated gradients; at the meta-level, both the task-specific and task-agnostic parts will be updated.  

\newcite{DBLPSunSZL19} and \newcite{DBLPGaoH0S19} studied few-shot relation classification problems. The dataset FewRel \cite{DBLPHanZYWYLS18} has totally 100 relation types. Some relations are used as training tasks, the remaining are dev/test tasks.

This kind of simulated few-shot study may have limited chance in reality, because for a test task with multiple few-shot classes, it is not easy to access highly related classes which have rich annotations and are in the same distribution as the few-shot classes. 

\paragraph{A domain is a task.} \newcite{DBLPuGYCPCTWZ18} and \newcite{DBLPGengLLZJS19} studied two text classification problems, each from multiple domains: one is multi-domain sentiment classification, the other is multi-domain intent classification. Each domain is treated as a task.  Then some domains are used as training tasks, some as test tasks. Technically, \newcite{DBLPuGYCPCTWZ18} proposed to use multiple metrics, learned in the training tasks, then a target few-shot task determines the best weighted combination from the set of learnt metrics.

Other works that conducted meta-learning within a specific problem include word sense disambiguation \cite{DBLP14355}, event detection \cite{DBLPDengZKZZC20,DBLPLaiDN20} and so on.

\subsection{Meta-learning on distinct problems}
To solve a target few-shot task, how to make use of other types of problems that have rich-annotation datasets is more challenging and practically useful.

\newcite{DBLPGuWCLC18}  framed low-resource translation as a meta-learning problem: eighteen high-resource language translation tasks as training tasks, five low-resource ones as testing tasks. Then, an extended MAML system is developed to handle this translation problem.

\newcite{DBLP03863} used GLUE \cite{DBngSMHLB19} tasks along with SNLI \cite{DBLwmanAPM15} as the training tasks, and evaluated on distinct tasks: entity typing, relation classification, sentiment classification, text categorization and SciTail (an entailment dataset by \cite{DBLPKhotSC18}). Their approach LEOPARD showed better performance than some competitive baselines including BERT \cite{DBLPinCLT19} fine-tuning, multi-task learning and prototypical networks.

\newcite{DBLPDouYA19} compared three typical optimization-based meta-learning, including MAML , First-order MAML and Reptile  on GLUE benchmarks: treating the low-resource tasks CoLA, MRPC, STS-B and RTE as the testing tasks, and the four high-resource tasks SST-2, QQP, MNLI and QNLI as the training tasks.  They also showed that meta-learning approaches surpass finetuned BERT and multi-task learning.

\section{Datasets for few-shot NLP}
\subsection{Class as task}

\paragraph{FewRel} \cite{DBLPHanZYWYLS18}, a relation classification dataset,  has 100 relations, each with 700 labeled sentences.  The official set uses 64, 16, and 20 relations as training/dev/test tasks. This dataset is constructed by manually annotating the distantly supervised  results on Wikipedia corpus and Wikidata  knowledge bases. Hence, all the training/dev/testing examples are from the same domain. A latest version, named FewRel-2.0 \cite{DBLPGaoHZLLSZ19}, added a new domain of test set and ``none-of-above'' relation. FewRel is reported by  \newcite{DBLPSunSZL19}, \newcite{DBLPGaoH0S19} and so on, and there is a leaderboard\footnote{\url{https://thunlp.github.io/1/fewrel1.html}}. 

\paragraph{SNIPS} \cite{DBLP10190}\footnote{\url{https://github.com/snipsco/nlu-benchmark/}} is an intent classification dataset with only seven intent types. Both \cite{DBLPXiaZYCY18,DBLP01881} used two intents as few-shot classes and  other  intents for training.

\subsection{Domain as task}

\paragraph{CLINC150} \cite{DBLPLarsonMPCLHKLLT19}\footnote{\url{https://github.com/clinc/oos-eval}} is an intent classification dataset. It has 23,700 instances in which 22,500 examples covers 150 intents, and 1,200 instances are out-of-scope. The 150 intents are  distributed in 10 domains: ``banking'', ``work'', ``meta'', ``auto \& commute'', ``travel'', ``home'', ``utility'', ``kitchen \& dining'', ``small talk'' and ``credit cards''; each domain has 15 intents.

\paragraph{ARSC} \cite{DBLPBlitzerDP07} is a sentiment classification dataset. It is comprised of Amazon reviews for 23 types of products (each product corresponds to a domain). For each product domain, there are three different binary classification tasks with different thresholds on the review rating: the tasks consider a review as positive if its rating = 5 stars, $>=$ 4 starts or $>=$ 2 stars \cite{DBLPuGYCPCTWZ18}. Therefore, this dataset has totally 23$\times$3=69 tasks. Both \cite{DBLPuGYCPCTWZ18,DBLP01907} used 12 tasks from 4 domains as target tasks, and the remaining domain tasks are training tasks.
 
Despite the existence of some few-shot NLP datasets, more challenging and realistic benchmarks are needed. As \newcite{DBLPTriantafillouZD20} claimed: (i) Real-life applications vary in  the numbers of classes and examples per class, and are unbalanced. Existing datasets, such as FewRel and CLINC150 mainly consider homogeneous learning tasks; (ii) We are eventually hoping that the model can generalize to tasks of new distributions, as \newcite{DBLP03863} did. Existing datasets often measure only within-dataset generalization. In terms of benchmark building, the computer vision community has made some progresses, such as the ``\textsc{META-DATASET}'' by \cite{DBLPTriantafillouZD20}.


\bibliography{emnlp2020.bib}

\begin{thebibliography}{34}
\expandafter\ifx\csname natexlab\endcsname\relax\def\natexlab#1{#1}\fi

\bibitem[{Bansal et~al.(2019)Bansal, Jha, and McCallum}]{DBLP03863}
Trapit Bansal, Rishikesh Jha, and Andrew McCallum. 2019.
\newblock Learning to few-shot learn across diverse natural language
  classification tasks.
\newblock \emph{CoRR}, abs/1911.03863.

\bibitem[{Blitzer et~al.(2007)Blitzer, Dredze, and Pereira}]{DBLPBlitzerDP07}
John Blitzer, Mark Dredze, and Fernando Pereira. 2007.
\newblock Biographies, bollywood, boom-boxes and blenders: Domain adaptation
  for sentiment classification.
\newblock In \emph{{ACL}}.

\bibitem[{Bowman et~al.(2015)Bowman, Angeli, Potts, and Manning}]{DBLwmanAPM15}
Samuel~R. Bowman, Gabor Angeli, Christopher Potts, and Christopher~D. Manning.
  2015.
\newblock A large annotated corpus for learning natural language inference.
\newblock In \emph{{EMNLP}}, pages 632--642.

\bibitem[{Coucke et~al.(2018)Coucke, Saade, Ball, Bluche, Caulier, Leroy,
  Doumouro, Gisselbrecht, Caltagirone, Lavril, Primet, and Dureau}]{DBLP10190}
Alice Coucke, Alaa Saade, Adrien Ball, Th{\'{e}}odore Bluche, Alexandre
  Caulier, David Leroy, Cl{\'{e}}ment Doumouro, Thibault Gisselbrecht,
  Francesco Caltagirone, Thibaut Lavril, Ma{\"{e}}l Primet, and Joseph Dureau.
  2018.
\newblock Snips voice platform: an embedded spoken language understanding
  system for private-by-design voice interfaces.
\newblock \emph{CoRR}, abs/1805.10190.

\bibitem[{Deng et~al.(2020)Deng, Zhang, Kang, Zhang, Zhang, and
  Chen}]{DBLPDengZKZZC20}
Shumin Deng, Ningyu Zhang, Jiaojian Kang, Yichi Zhang, Wei Zhang, and Huajun
  Chen. 2020.
\newblock Meta-learning with dynamic-memory-based prototypical network for
  few-shot event detection.
\newblock In \emph{{WSDM}}, pages 151--159.

\bibitem[{Devlin et~al.(2019)Devlin, Chang, Lee, and Toutanova}]{DBLPinCLT19}
Jacob Devlin, Ming{-}Wei Chang, Kenton Lee, and Kristina Toutanova. 2019.
\newblock {BERT:} pre-training of deep bidirectional transformers for language
  understanding.
\newblock In \emph{{NAACL-HLT}}, pages 4171--4186.

\bibitem[{Dou et~al.(2019)Dou, Yu, and Anastasopoulos}]{DBLPDouYA19}
Zi{-}Yi Dou, Keyi Yu, and Antonios Anastasopoulos. 2019.
\newblock Investigating meta-learning algorithms for low-resource natural
  language understanding tasks.
\newblock In \emph{{EMNLP-IJCNLP}}, pages 1192--1197.

\bibitem[{Finn et~al.(2017)Finn, Abbeel, and Levine}]{DBLPFinnAL17}
Chelsea Finn, Pieter Abbeel, and Sergey Levine. 2017.
\newblock Model-agnostic meta-learning for fast adaptation of deep networks.
\newblock In \emph{{ICML}}, pages 1126--1135.

\bibitem[{Gao et~al.(2019{\natexlab{a}})Gao, Han, Liu, and Sun}]{DBLPGaoH0S19}
Tianyu Gao, Xu~Han, Zhiyuan Liu, and Maosong Sun. 2019{\natexlab{a}}.
\newblock Hybrid attention-based prototypical networks for noisy few-shot
  relation classification.
\newblock In \emph{{AAAI}}, pages 6407--6414.

\bibitem[{Gao et~al.(2019{\natexlab{b}})Gao, Han, Zhu, Liu, Li, Sun, and
  Zhou}]{DBLPGaoHZLLSZ19}
Tianyu Gao, Xu~Han, Hao Zhu, Zhiyuan Liu, Peng Li, Maosong Sun, and Jie Zhou.
  2019{\natexlab{b}}.
\newblock Fewrel 2.0: Towards more challenging few-shot relation
  classification.
\newblock In \emph{{EMNLP-IJCNLP}}, pages 6249--6254.

\bibitem[{Geng et~al.(2019)Geng, Li, Li, Zhu, Jian, and Sun}]{DBLPGengLLZJS19}
Ruiying Geng, Binhua Li, Yongbin Li, Xiaodan Zhu, Ping Jian, and Jian Sun.
  2019.
\newblock Induction networks for few-shot text classification.
\newblock In \emph{{EMNLP-IJCNLP}}, pages 3902--3911.

\bibitem[{Gu et~al.(2018)Gu, Wang, Chen, Li, and Cho}]{DBLPGuWCLC18}
Jiatao Gu, Yong Wang, Yun Chen, Victor O.~K. Li, and Kyunghyun Cho. 2018.
\newblock Meta-learning for low-resource neural machine translation.
\newblock In \emph{{EMNLP}}, pages 3622--3631.

\bibitem[{Han et~al.(2018)Han, Zhu, Yu, Wang, Yao, Liu, and
  Sun}]{DBLPHanZYWYLS18}
Xu~Han, Hao Zhu, Pengfei Yu, Ziyun Wang, Yuan Yao, Zhiyuan Liu, and Maosong
  Sun. 2018.
\newblock {FewRel}: {A} large-scale supervised few-shot relation classification
  dataset with state-of-the-art evaluation.
\newblock In \emph{{EMNLP}}, pages 4803--4809.

\bibitem[{Holla et~al.(2020)Holla, Mishra, Yannakoudakis, and
  Shutova}]{DBLP14355}
Nithin Holla, Pushkar Mishra, Helen Yannakoudakis, and Ekaterina Shutova. 2020.
\newblock Learning to learn to disambiguate: Meta-learning for few-shot word
  sense disambiguation.
\newblock \emph{CoRR}, abs/2004.14355.

\bibitem[{Hospedales et~al.(2020)Hospedales, Antoniou, Micaelli, and
  Storkey}]{DBLP05439}
Timothy~M. Hospedales, Antreas Antoniou, Paul Micaelli, and Amos~J. Storkey.
  2020.
\newblock Meta-learning in neural networks: {A} survey.
\newblock \emph{CoRR}, abs/2004.05439.

\bibitem[{Jiang et~al.(2018)Jiang, Havaei, Chartrand, Chouaib, Vincent, Jesson,
  Chapados, and Matwin}]{jiang2018attentive}
Xiang Jiang, Mohammad Havaei, Gabriel Chartrand, Hassan Chouaib, Thomas
  Vincent, Andrew Jesson, Nicolas Chapados, and Stan Matwin. 2018.
\newblock Attentive task-agnostic meta-learning for few-shot text
  classification.
\newblock In \emph{NeurIPS Meta-Learning Workshop}.

\bibitem[{Khot et~al.(2018)Khot, Sabharwal, and Clark}]{DBLPKhotSC18}
Tushar Khot, Ashish Sabharwal, and Peter Clark. 2018.
\newblock {SciTaiL}: {A} textual entailment dataset from science question
  answering.
\newblock In \emph{{AAAI}}, pages 5189--5197.

\bibitem[{Koch et~al.(2015)Koch, Zemel, and Salakhutdinov}]{koch2015siamese}
Gregory Koch, Richard Zemel, and Ruslan Salakhutdinov. 2015.
\newblock Siamese neural networks for one-shot image recognition.
\newblock In \emph{ICML deep learning workshop}, volume~2.

\bibitem[{Lai et~al.(2020)Lai, Dernoncourt, and Nguyen}]{DBLPLaiDN20}
Viet~Dac Lai, Franck Dernoncourt, and Thien~Huu Nguyen. 2020.
\newblock Exploiting the matching information in the support set for few shot
  event classification.
\newblock In \emph{{PAKDD}}, pages 233--245.

\bibitem[{Larson et~al.(2019)Larson, Mahendran, Peper, Clarke, Lee, Hill,
  Kummerfeld, Leach, Laurenzano, Tang, and Mars}]{DBLPLarsonMPCLHKLLT19}
Stefan Larson, Anish Mahendran, Joseph~J. Peper, Christopher Clarke, Andrew
  Lee, Parker Hill, Jonathan~K. Kummerfeld, Kevin Leach, Michael~A. Laurenzano,
  Lingjia Tang, and Jason Mars. 2019.
\newblock An evaluation dataset for intent classification and out-of-scope
  prediction.
\newblock In \emph{{EMNLP-IJCNLP}}, pages 1311--1316.

\bibitem[{Nichol et~al.(2018)Nichol, Achiam, and Schulman}]{DBLP02999}
Alex Nichol, Joshua Achiam, and John Schulman. 2018.
\newblock On first-order meta-learning algorithms.
\newblock \emph{CoRR}, abs/1803.02999.

\bibitem[{Snell et~al.(2017)Snell, Swersky, and Zemel}]{DBLPSnellSZ17}
Jake Snell, Kevin Swersky, and Richard~S. Zemel. 2017.
\newblock Prototypical networks for few-shot learning.
\newblock In \emph{{NeurIPS}}, pages 4077--4087.

\bibitem[{Sui et~al.(2020)Sui, Chen, Mao, Qiu, Liu, and Zhao}]{DBLP01907}
Dianbo Sui, Yubo Chen, Binjie Mao, Delai Qiu, Kang Liu, and Jun Zhao. 2020.
\newblock Knowledge guided metric learning for few-shot text classification.
\newblock \emph{CoRR}, abs/2004.01907.

\bibitem[{Sun et~al.(2019)Sun, Sun, Zhou, and Lv}]{DBLPSunSZL19}
Shengli Sun, Qingfeng Sun, Kevin Zhou, and Tengchao Lv. 2019.
\newblock Hierarchical attention prototypical networks for few-shot text
  classification.
\newblock In \emph{{EMNLP-IJCNLP}}, pages 476--485.

\bibitem[{Sung et~al.(2018)Sung, Yang, Zhang, Xiang, Torr, and
  Hospedales}]{DBLPSungYZXTH18}
Flood Sung, Yongxin Yang, Li~Zhang, Tao Xiang, Philip H.~S. Torr, and
  Timothy~M. Hospedales. 2018.
\newblock Learning to compare: Relation network for few-shot learning.
\newblock In \emph{{CVPR}}, pages 1199--1208.

\bibitem[{Thrun and Pratt(1998)}]{DBLPThrunP98}
Sebastian Thrun and Lorien~Y. Pratt. 1998.
\newblock Learning to learn: Introduction and overview.
\newblock In \emph{Learning to Learn}, pages 3--17.

\bibitem[{Triantafillou et~al.(2020)Triantafillou, Zhu, Dumoulin, Lamblin,
  Evci, Xu, Goroshin, Gelada, Swersky, Manzagol, and
  Larochelle}]{DBLPTriantafillouZD20}
Eleni Triantafillou, Tyler Zhu, Vincent Dumoulin, Pascal Lamblin, Utku Evci,
  Kelvin Xu, Ross Goroshin, Carles Gelada, Kevin Swersky, Pierre{-}Antoine
  Manzagol, and Hugo Larochelle. 2020.
\newblock Meta-dataset: {A} dataset of datasets for learning to learn from few
  examples.
\newblock In \emph{{ICLR}}.

\bibitem[{Vanschoren(2018)}]{DBLPabs03548}
Joaquin Vanschoren. 2018.
\newblock Meta-learning: {A} survey.
\newblock \emph{CoRR}, abs/1810.03548.

\bibitem[{Vilalta and Drissi(2002)}]{DBLPVilaltaD02}
Ricardo Vilalta and Youssef Drissi. 2002.
\newblock A perspective view and survey of meta-learning.
\newblock \emph{Artif. Intell. Rev.}, 18(2):77--95.

\bibitem[{Vinyals et~al.(2016)Vinyals, Blundell, Lillicrap, Kavukcuoglu, and
  Wierstra}]{DBLPVinyalsBLKW16}
Oriol Vinyals, Charles Blundell, Tim Lillicrap, Koray Kavukcuoglu, and Daan
  Wierstra. 2016.
\newblock Matching networks for one shot learning.
\newblock In \emph{{NeurIPS}}, pages 3630--3638.

\bibitem[{Wang et~al.(2019)Wang, Singh, Michael, Hill, Levy, and
  Bowman}]{DBngSMHLB19}
Alex Wang, Amanpreet Singh, Julian Michael, Felix Hill, Omer Levy, and
  Samuel~R. Bowman. 2019.
\newblock {GLUE:} {A} multi-task benchmark and analysis platform for natural
  language understanding.
\newblock In \emph{{ICLR}}.

\bibitem[{Xia et~al.(2020)Xia, Zhang, Nguyen, Zhang, and Yu}]{DBLP01881}
Congying Xia, Chenwei Zhang, Hoang Nguyen, Jiawei Zhang, and Philip~S. Yu.
  2020.
\newblock {CG-BERT:} conditional text generation with {BERT} for generalized
  few-shot intent detection.
\newblock \emph{CoRR}, abs/2004.01881.

\bibitem[{Xia et~al.(2018)Xia, Zhang, Yan, Chang, and Yu}]{DBLPXiaZYCY18}
Congying Xia, Chenwei Zhang, Xiaohui Yan, Yi~Chang, and Philip~S. Yu. 2018.
\newblock Zero-shot user intent detection via capsule neural networks.
\newblock In \emph{{EMNLP}}, pages 3090--3099.

\bibitem[{Yu et~al.(2018)Yu, Guo, Yi, Chang, Potdar, Cheng, Tesauro, Wang, and
  Zhou}]{DBLPuGYCPCTWZ18}
Mo~Yu, Xiaoxiao Guo, Jinfeng Yi, Shiyu Chang, Saloni Potdar, Yu~Cheng, Gerald
  Tesauro, Haoyu Wang, and Bowen Zhou. 2018.
\newblock Diverse few-shot text classification with multiple metrics.
\newblock In \emph{{NAACL-HLT}}, pages 1206--1215.

\end{thebibliography}
\bibliographystyle{acl_natbib}

\end{document}